# A computational model and tool for generating more novel opportunities in professional innovation processes

## Authors
Neil Maiden, Konstantinos Zachos, James Lockerbie, Kostas Petrianakis*, Amanda Brown

## Affiliation
City St George's, University of London, except *

## Corresponding author
Neil Maiden, Neil.Maiden.1@citystgeorges.ac.uk

## Funding
The development of INSIGHTS was funded by UKRI's Research England Development fund

## Abstract
This paper presents a new computational model of creative outcomes, informed by creativity theories and techniques, which was implemented tool to generate more novel opportunities for innovation projects. The model implemented five functions that were developed to contribute to the generation of innovation opportunities with higher novelty without loss of usefulness. The model was evaluated using opportunities generated for an innovation project in the hospitality sector. The evaluation revealed that the computational model generated outcomes that were more novel and/or useful than outcomes from Notebook LM and ChatGPT4o. However, not all of the model's functions contributed to the generation of more novel opportunities, leading to new directions for further model development.

# 1. Introduction

The evidence that generative AI technologies are impacting creative professional work is growing. E.g., management consultants have used ChatGPT to explore larger problem and solution spaces and generate more novel outcomes with which to innovate [1] and architects reported that Midjourney use enhanced their divergent and convergent creative processes [2]. However, generative AI's reliance on pre-existing data and probabilistic language modelling to generate content often limits it to more incremental forms of creative outcomes (e.g., [3]). Questions remain about its capabilities to explore new spaces of outcomes that are more novel as well as useful (e.g., [4]).

What is more, generative AI does not reason systematically with codified knowledge about creative outcomes – the types of learned knowledge that people apply when thinking creatively. Generative AI and people arrive at creative outcomes in very different ways [3]. Unlike chatbots that typically generate sets of ideas directly in response to user prompts, creative thinking techniques such as constraint removal (e.g., [5]) direct people through a series of steps to define the problem, open up new spaces of possibilities then discover more novel outcomes in these new spaces. We argue that this failure to manage different spaces of possibilities limits generative AI's ability to produce more creative outcomes. Explicit representation of and control over these spaces and the discovery of opportunities in them offers greater control over this creative thinking, consistent with human-centred AI principles [6].

The paper describes how to generate outcomes in innovation projects that are more creative than with existing generative AI tools. Creative outcomes are both novel and useful [7], but this definition does not distinguish between different degrees of novelty. Therefore, our research draws on existing creative theories (e.g., [8]) and techniques [9] to define how to discover more novel outcomes using rules with which to explore known spaces of outcomes. Two forms of a new model were developed. The first, a descriptive model, defines the distinguishing properties of creative outcomes as a set of five functions with which to generate them, derived from existing creativity theories and techniques, which are a major advance over the previous version of the model [10]. The second, a computational implementation of the descriptive model, integrates different AI technologies to generate creative outcomes to implement the five functions. This computational model was then evaluated to compare the relative novelty and usefulness of outcomes generated with settings that control different model functions, then with the novelty and usefulness of outcomes generated by two other tools – ChatGPT and Notebook LM – in a test innovation consulting domain.

# 2. Literature review

Numerous theories of creative processes and outcomes exist. Most define creative outcomes as both novel and useful [7], but most creativity research investigates novelty (e.g., [11] [12]) rather than usefulness. Shneiderman [13] distinguishes between theories that are inspirationalist, situationalist or structuralist. Structuralist theories emerged from theories of information processing and describe creative possibilities in terms of the information manipulated to generate them. This focus on information about possibilities provided the baseline upon which the new model of creative outcomes was constructed.

Structuralist theories often define creative problem solving as iterations of divergent then convergent possibilities to explore spaces of larger numbers of less complete possibilities then fewer but more complete ones [14]. Within this framing, Boden [8] distinguished between two types of creativity – exploratory and transformational. Exploratory creativity assumes a defined conceptual space of partial and complete possibilities to explore – a space that also implies the existence of rules that define the space. Changes to these rules produce what might be thought of as a paradigm shift, called transformational creativity [8]. Possibilities that are novel and useful are reached in each space by what are called generative rules [8]. Different rules discover more divergent and convergent possibilities and are often operationalised as guidance within creative thinking techniques such as SCAMPER [15]. One specific form of exploratory creativity, called combinational creativity, makes unfamiliar connections between familiar items in the pre-defined search space, using a different set of generative rules [8]. Early computerised models of transformational, exploratory and combinational creativity have been reported (e.g., [16]) but did not integrate codified knowledge from diverse creativity theories and techniques, or from their application in domains such as innovation consulting.

Building on different types of conceptual space, creative possibilities have also been defined in the context of a need not fulfilled by existing solutions in the same class [17] [18]. An incomplete set of known solutions for a class can define the space of yet-to-be-discovered partial and complete possibilities for it, and the relative novelty of each possibility can be measured by how different it is from the set of known possibilities, often measured by the distance between it and the centroid of the nearest cluster of possibilities for that class [18], using e.g., semantic networks [19]. Returning to Boden [8], discovering possibilities furthest from the centroid of each conceptual space defined by a class can result in more novel creative outcomes using an exploratory creative thinking style.

Creative possibilities have also been observed to have qualities that can repeat across domains. E.g., the TRIZ creative thinking technique codified repeating solution patterns for solving mechanical engineering problems – principles such as *segmentation* and *do it in reverse* that were available to reuse when solving problems [20]. An equivalent set of principles was developed methodologically to describe the repeating creative qualities of digital services [9] – qualities such as *trust* and *playful*. Although these TRIZ principles and trigger qualities have the potential to direct computational algorithms to discover more novel possibilities in conceptual spaces, no such applications have yet been reported.

Generative Pre-trained Transformer (GPT) neural networks seek to predict accurate sequences of words for input text prompts using Large Language Models (LLMs) trained on massive language datasets. Implementations such as GPT5 [21] and Claude-3 [22] can create possibilities by manipulating large volumes of information from beyond organisational boundaries. Chatbot interactions with these models in business contexts have been shown to support consultants to generate more ideas (e.g., [23]) and higher quality results (e.g., [1]).

However, generative AI's reliance on pre-existing data and probabilistic language modelling can limit it to incrementally creative outcomes (e.g., [3]). Questions remain about GPT's capabilities to explore spaces of more novel possibilities (e.g., [4]). These questions are supported by empirical findings. E.g., although generative AI use enhanced creative thinking by offering new possibilities [24] and exploring larger ideas spaces [23], it also reduced the collective novelty of these possibilities [24] and led to higher fixation on examples as well as fewer solutions with lower originality compared to a baseline [25]. Indeed, generative AI ideas often trade-off novelty for usefulness [26], and little evidence for generative AI producing more novel outcomes has been published.

## 3. The new descriptive model of creative outcomes

The new descriptive model integrates characteristics of creative outcomes from different creativity theories and techniques. It serves three purposes. The first was to define more novel and useful outcomes in the context of innovation processes that manipulate large volumes of information about problems and possible solutions. The decision to focus on innovation consulting was deliberate. Grounding the model in a concrete domain informed its development and enabled more effective evaluations of it, consistent with our design science approach [27]. The second and third purposes were to define the characteristics of more novel and useful outcomes with sufficient precision to enable the design and implement a computational version of the model, and to link the computational model with existing creativity theories and techniques to provide a rationale for it.

The model integrates five functions. Each manipulates information about innovation problems and solutions to contribute to the generation of creative outcomes in the structuralist tradition [13], consistent with exploratory and combinational creative thinking [8]. Each was developed to generate more complex opportunities rather than simpler ideas, to provide more support for innovation processes. E.g., generated business opportunities describe not only the idea but also potential markets for that idea, and policy opportunities also describe its positive contribution to target audiences and their lives.

Each of the five functions was derived from a different characteristic of creative opportunities and the conceptual spaces in which these opportunities can be discovered:

1. **Generative-rules**: more creative opportunities can be discovered from systematic explorations of opportunity spaces composed of large numbers of possible opportunities. Boden [8] describes how problem and solution information can be manipulated systematically using generative rules to discover opportunities that are creative from the very large numbers of the opportunities possible, consistent with the exploratory style of creative thinking. Tried-and-tested creative thinking techniques (e.g., [28] [29] [30]) also direct their users to apply different generative rules – rules that can be codified in the model to discover opportunities more systematically.
2. **Pivot-opportunities**: more creative opportunities often emerge iteratively from processes composed of multiple steps that increase the novelty of opportunities (e.g., [31] [32]). In each step, the outcome from one generative rule becomes the input to the next, to systematically explore more areas of an opportunity space. This "pivoting" on discovered opportunities can contribute to more complete explorations of spaces as well as overcome idea fixation in people [33].
3. **Classes-of-needs**: opportunity spaces can describe all of the possible opportunities that meet the needs for recognised classes [18]. Therefore, discovering and defining these spaces using available information about existing needs and their solutions can direct the effective and systematic discovery of creative opportunities using generative rules in such space.
4. **Atypical-opportunities**: more creative opportunities in each space have higher semantic distances from the centroid of that space [18]. Novelty is an essential attribute of any creative outcome [7], and more novel opportunities have greater semantic distances from the most common examples of these opportunities [18]. Systematically applying generative rules to discover opportunities that have greater differences with common examples has the potential to direct creative thinking to discover more novel yet still novel opportunities.
5. **Creative-qualities**: more creative opportunities have predefined qualities. Whereas the first four functions define the creative potential of opportunities relative to others, the fifth identifies qualities of single opportunities associated with creative outcomes, e.g., digital products have recurring qualities (e.g., more *informative* or *playful*) that render them more creative in the eyes of users [9], similar to earlier taxonomies of recurring qualities of creative engineering solutions [20]. Therefore, generative rules can also be applied systematically to discover opportunities with these qualities.

Consider an example of an innovation consultant with expertise in biophilic buildings exploring creative opportunities to redevelop a pub and its grounds. The model describes different spaces of opportunities using biophilic design information about, e.g., technologies and case studies. Figure 1 depicts three opportunity spaces uncovered from this information, each bounded by dotted lines. Large numbers of opportunities depicted as black dots can be discovered via systematic explorations. Each opportunity has a different semantic distance d to its centroid – the white dot.

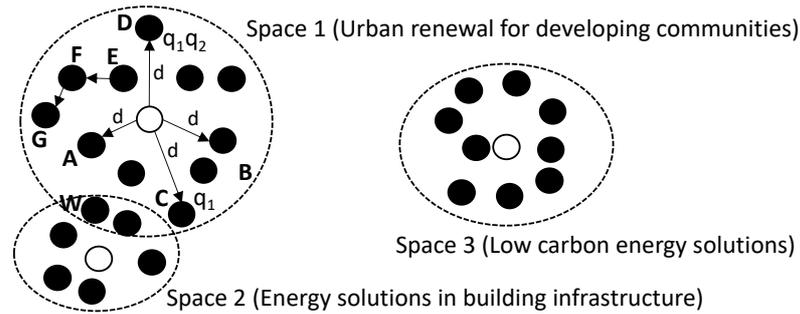

*Figure 1: A visual representation of three spaces of opportunities for biophilic public buildings discovered from problem and solution information about such buildings: 1) Urban renewal for developing communities; 2) Energy solutions in building infrastructures, and 3) Low carbon energy solutions. Each space's centroid describes the most average and hence the least novel opportunity in that space, and all other opportunities in each space have a semantic distance from this average opportunity.*

Four of the opportunities in Space 1 (*Urban renewal for developing communities*) are labelled A, B, C and D. Opportunity A has a shorter distance to the centroid of the space than do B, C and D, so A is the least novel, and hence is the least creative of the four. One such opportunity might be *to turn the pub into a community co-working cafe*. Opportunities B, C and D all have larger but similar semantic distances from the centroid, so according to the model's **atypical-opportunities** function all have higher potential novelty than A. While B (e.g., *an art and nature fusion hub*) and C (e.g., *a digital detox retreat*) each have one predefined creative quality ($q_1$ (*more ecological*) and $q_2$ (*healthier*)) [9], D (e.g., *bio integrated health trails based out of the pub*) has both, so D has the potential to be more novel than B or C according to the model's **creative-qualities** function. Many opportunities in each space can be discovered using a single generative rule, however more novel opportunities with greater semantic distances from centroids can be discovered by pivoting on discovered opportunities using the **pivot-opportunities** function. E.g., Figure 1 depicts more novel Opportunity G being discovered from a pivot on Opportunity F using a generative rule which, in turn, was discovered from a pivot on Opportunity E using a different generative rule.

On its own, the model's contribution to discovering creative outcomes in innovation consulting projects is limited. Therefore, a computational version of the model was also developed to investigate empirically whether it can generate novel and useful outcomes using information typically available to innovation consulting projects.

## 4. The computational model of creative outcomes

A software implementation of the descriptive model was developed to generate opportunities systematically in spaces of opportunities uncovered from innovation project information. The implementation codifies each of the descriptive model's five functions as part of an automated four-step procedure depicted graphically in Figure 2. The procedure assumes interventions by innovation consultants during the fourth stage, to select between large number of candidate opportunity spaces and then the opportunities, to direct the generation process. The software implementation combined different types of technologies including but not limited to generative AI.

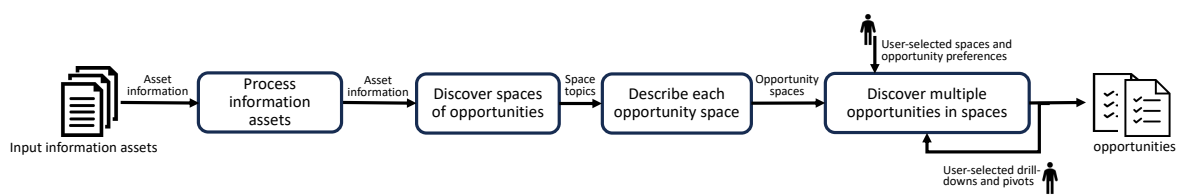

*Figure 2. The four steps of the procedure implemented by the computational model of creative outcomes.*

The input to the computational model is a set of information assets related to one innovation project. The model assumes these assets describe large volumes of unstructured information about the innovation problem and possible solutions that can be stored as PDF, MS Word and PowerPoint file types, include tables or graphics, and be expressed in different languages. Information describing one space of opportunities can be spread across files, repeating with different degrees of variance, and interleaved with information describing other spaces. The first step of the procedure extracts this unstructured information for processing. The second automatically discovers spaces of opportunities from the processed inputs using the model's **class-of-needs** function. The third automatically generates descriptions of these spaces to guide consultants to decide which to explore, also as part of the **class-of-needs** function. The fourth step automatically applies bespoke generative rules to generate candidate opportunities drawing on the model's **generative-rules**, **pivot-opportunities**, **atypical-opportunities** and **creative-qualities** functions. Each procedure step is described in more detail.

### 4.1. Step 1: Process information assets

The procedure extracts text from each file type (e.g., from PDF, HTML and image files using optical character recognition) with different python libraries, exploring nested pockets of content elements to retrieve all text. The text extraction libraries are *pdftotext* and *pypdf* for PDF files, *pytesseract* and *pdf2image* for OCR recognition, *docx2python* for MS Word files, *python-ppt* for MS PowerPoint files and *html2text* for HTML files. A cleaning algorithm manipulates all of the text to remove extra spaces, standardise characters and expunge anomalies. If text was written in languages other than English, it is translated using the *DeepL* service [34].

### 4.2. Step 2: Discover spaces of opportunities

The procedure then applies topic modelling to the extracted, cleaned and translated text to generate opportunity spaces. Topic modelling is a form of statistical modelling that uses unsupervised machine learning to discover clusters of similar words within the extracted text corpus. The *BERTopic* software was chosen because of its reliability and scalability with incomplete and unstructured information. It implements UMAP to reduce the dimensionality of text embeddings, HDBSCAN to cluster the reduced embeddings, then class-based TF-IDF to extract and reduce topics, and Maximal Marginal Relevance to improve the coherence of words. As a consequence, semantic structures in the text are used to cluster unstructured data without predefined tags, training data or user intervention. Manually set parameter values reflect the file sizes, numbers and content types. Each generated cluster defines one opportunity space as a multi-dimensional space of information pieces expressed as ten topic terms, in which more similar pieces are closer together. These topic terms, in descending order of their weightings, define the subject matter of opportunities in each space. E.g., the top ten topic terms ordered by their weightings that describe the *urban renewal and community development* opportunity space from Figure 1 are: *project*, *development*, *urban*, *nature*, *community*, *spaces*, *design*, *phoenix*, *human* and *social*. Each generated set of opportunity spaces is ordered according to the number of different clustered topic terms in each space, indicating each space's coverage of the information extracted from the information assets.

To increase choice for opportunity generation, this step generates three sets of opportunity spaces of different sizes for each set of input assets – between four and eight broad opportunity spaces, eight and 15 typical opportunity spaces, and 15 and 30 narrow opportunity spaces. The sizes of these sets were based on feedback on model outcomes from innovation consultants. Returning to our biophilic design example, narrow opportunity spaces include *urban renewal and community development* and *energy solutions in* *building infrastructure*. Broad ones include *sustainable urban development* and *platform resilience in local systems*.

### 4.3 Step 3: Describe each opportunity space

To generate natural language descriptions of each opportunity space for innovation consultants to choose from, the ordered topic terms are input into an operation that invokes an API call to GPT4o to generate these natural language descriptions. The output from this simple step is a short label and 100-word description of each space and its distinguishing characteristics. The word length was set during user testing with innovation consultants to balance between readability and detail. The GPT prompt called by the operation is depicted in Figure 3.

> **GPT settings were temperature=0.7, top_p=1, frequence penalty=0, presence penalty=0.**
>
> You are an experienced consultant. Generate a single paragraph of 100-words that accurately describes the area characterised by the following terms. Also generate a short name for the summary that does not contain a verb. Do not generate other content. The summary should use direct language that can be understood by someone with limited design knowledge. The first sentence should start with the expression "This area is". The terms are: {**Topic terms**}.

*Figure 3. The automated procedure's GPT prompt to create natural language description of one discovered opportunity space.*

### 4.4. Step 4: Discover multiple opportunities in spaces

Generative rules are applied to discover opportunities in selected spaces. Different rules implement different combinations of the **generative-rules** and **pivot-opportunities** functions. Each rule is codified as a bespoke parameterised operation that manipulates multiple inputs to populate a bespoke GPT4o prompt via an API and generate ten opportunities. Different rules discover different types of opportunities. Business-type opportunities build on ideas that can support organisations to innovate, policy-type opportunities build on ideas that policy makers can implement to improve people's lives, and technical design opportunities build on ideas related to the development and application of different technologies. An example of one of these generative rules – to discover business-type opportunities in a single space – is presented in Figure 4, with input variables shown in bold.

> **GPT settings were top_p=1, frequence penalty=0, presence penalty=0.**
>
> As an experienced business analyst/consultant, recommend 10 business innovations. Each innovation must be different. The idea underpinning each innovation must centre on one of more topics from {**Selected space topic terms**} and incorporate interpretations of other themes referenced in the following {**100-word description of topic**}. Each idea must be {**Novelty setting**} compared to other possible innovations. Each opportunity should align with the following needs: {**Custom text**}. Describe each innovation in terms of its underlying idea, beneficiaries and markets, expected outcomes, and possible implementation strategy. Each innovation should be described in at least 100 words and presented as a title of up to 10 words, a colon character and the 100-word description. Don't use bullets, and new lines either in the title or the description. {**Selected creative qualities**}.

*Figure 4. The automated procedure's GPT prompt to generate business-type opportunities in one selected opportunity space using a selected novelty setting, custom text entered by a user, and creative qualities also selected by a user.*

The different input variables that control the model's functions are:

- {**Novelty setting**} and {**Selected space topic terms**} together control how unusual or otherwise new opportunities should be – their semantic distances to the centroid of the space – as part of the **atypical-opportunities** function. The novelty setting defines five predefined option values from *very prototypical* (i.e., close to the centroid) to *highly unusual* (far from the centroid) that each rule uses to select between the ten ordered topic terms that describe the opportunity space and instantiate the selected space topic terms. E.g., for *very prototypical* opportunities the rule selects five topic terms with the highest weightings, whereas for *highly unusual opportunities* it selects three with the lowest weightings. These terms are associated with different parameterised GPT temperature settings also associated with each novelty setting.
- {**Selected creative qualities**} control the model's **creative-qualities** function. Each rule accepts up to three of 22 qualities associated with digital products and services [Giunta et al. 2022]: *increased service, added information choice, greater participation, more connected, greater trust, more convenient, greener, more entertaining, more durable, cheaper to run, more adaptable, more informative, more fashionable, inspirational, higher productivity, greater independence, more playful, more beautiful, more direct, healthier, more influential*, and *younger*.
- {**Custom text**} is an input text string similar to a prompt with generative AI chatbots. It directs the rule to align generated opportunities with the input text's content, written directly by innovation consultants and/or can be automatically generated by summarizing document files uploaded into the tool by the consultants.

A third set of rules were developed to implement the model's **pivot-opportunities** function. Each pivot uses generated and selected opportunities to generate new ones. Each rule is similar to those described in Figure 5, but the inputs are extracted from the selected opportunity rather than topic.

---

**GPT settings top_p=1, frequence penalty=0, presence penalty=0.**

As an experienced policy maker, recommend 10 specific alternative policy opportunities with considerable novelty that are creative variations of the following idea. {**Selected opportunity**}. All 10 variations must be within the space described by one of two different policy areas and their topics. The first, third and fifth should be from the first policy area, the second, fourth, sixth should be from the second policy area, and the remaining four opportunities should integrate opportunities from both policy areas. The first policy area is as follows. {**100-word description of first topic**}. The topics in the area are {**First space topic terms**}. The second policy area is as follows. {**100-word description of second topic**}. The topics in the area are {**Second space topic terms**}. Each policy opportunity should be based on at least one of the topics. Each opportunity should be {**Novelty settings**} of those that are possible in each policy area. Each opportunity should align with the following needs: {**Custom text**}. Describe each opportunity to define how it is focused on its audience, responsive and accountable to this audience, effective and efficient, and improving lives. Each policy should be described in at least 120 words and presented as a title of up to 10 words, a colon character and the 120-word description. Don't use bullets, and new lines either in the title or the description. {**Selected creative qualities**}

---

*Figure 5. The automated procedure's GPT prompt to generate policy-type opportunities in one selected opportunity space using one selected opportunity, a selected novelty setting, custom text entered by a user, creative qualities also selected by a user.*

This version of the computational model was integrated into an interactive tool called INSIGHTS that innovation consultants can use to control the model's stages and functions. More information about the INSIGHTS tool is reported in Maiden et al. [2025]. The remainder of this paper reports an evaluation of the computational model.

## 5. Evaluating the computational model of creative outcomes

Previous versions of the computational model and INSIGHTS tool had been evaluated using qualitative data (Maiden et al. 2025). E.g., engineers responsible for dishwasher innovation used it during workshops to explore new opportunities for product releases. Powerplant employees also used it to explore opportunities for biomass biproducts and found it to be effective for exploring new areas and generating more novel opportunities. However, no systematic analysis of opportunities generated by the computational model's functions had been undertaken. Therefore, a controlled comparison of model outcomes in response to different inputs to explore hypothesized contributions of the model versus LLM-based tools and the model's five functions was undertaken.

## 5.1. The evaluation project and baseline opportunity request

A new project was set up with nine information assets – PDF documents written in English and reporting different aspects of UK and related hospitality sectors. Each asset was publicly available and discovered using a Google search. The shortest was six pages long, the longest 55 pages, see Table 1. The computational model's procedure processed these assets to generate broad, typical and narrow opportunity spaces. Professionals working in two UK hotels generated multiple opportunities in these spaces with the INSIGHTS tool that demonstrated the relevance of the assets to the hospitality sector.

| Information asset title | Information asset author | Pages |
|---|---|---|
| Coronavirus and its impact on UK hospitality | UK Office for National Statistics | 15 |
| The European hospitality industry outlook | Deloitte | 12 |
| The economic contribution of the UK hospitality industry | Ignite Economics | 25 |
| UK hospitality's next challenge: How sourcing the right people could net a £36bn bonus for the sector | UK Government: Department of Business and Trade | 33 |
| Hospitality strategy: reopening, recovery, resilience | Barclays Bank | 31 |
| Hospitality and Tourism workforce landscape | Economic Insight | 55 |
| OPUS Report: UK Hospitality Market | OPUS Business Advisory Group | 6 |
| The Digital Skills Gap – Is it Time to Rethink the Needs of Tourism and Hospitality Organizations in the UK? | Journal of Hospitality & Tourism Education | 12 |
| UK Hospitality Workforce Commission 2030 report: The changing face of hospitality | UK Hospitality | 16 |

Table 1. The titles, author organization and page lengths of nine information assets used to generate the INSIGHTS UK hospitality sector project.

Next, INSIGHTS requests were run to generate baseline opportunities for the evaluation. Each run was repeated three times to generate opportunities for each of three narrow opportunity spaces uncovered from the information assets: *Commercial property management support* [the large opportunity space], *Revitalizing local streets* [the fifth largest space], and *Hospitality recovery and resilience* [the smallest space]. In turn, each of these requests was run three times to generate policy opportunities, then business opportunities, then technical opportunities, using the input values in Table 2. To generate 30 opportunities for each run, 10 were generated directly from the topic terms of the opportunity space, 10 were generated from pivoting on the first of the opportunities in that space, and 10 were generated from pivoting on one of subsequent opportunities, again within that space.

| INSIGHTS request attribute | Entered value |
|---|---|
| Opportunities type | Policy / Business / Technical design |
| Describe a specific idea you want to explore | Develop an innovative [policy/business/technical design] opportunity that support seaside towns to regenerate by attracting new investment linked to new areas of growth |
| How unusual or otherwise do you want the new opportunities to be? | Generate highly unusual opportunities |
| Do you want the opportunities to have any of the following creative qualities? | Without creative qualities or mega-trends |

Table 2. Values input to the computational model of creative outcomes via the INSIGHTS tool to generate the baseline opportunities for the UK hospitality project.

The outcome from these nine baseline requests was a set of 270 innovation opportunities – 90 policy opportunities, 90 business opportunities, and 90 technical design opportunities. These INSIGHTS opportunities provided the baseline for two comparisons – one that compared them to opportunities generated by two other generative AI tools, the other that compared the opportunities generating using different model functions.

## 5.2 Comparing INSIGHTS to Notebook LM and ChatGPT

The 270 baseline INSIGHTS opportunities were compared to opportunities generated by two general-purpose generative AI tools often used for idea generation in innovation and consulting work. The first was Google's Notebook LM, a tool that grounds language models in user-provided documents to create a personalized LLM [35]. The comparison to INSIGHTS was obvious – both generated guidance including ideas based on information extracted from a set of uploaded documents. Therefore, a new project was set up on Notebook LM and populated with the same nine information assets. Then, to generate opportunities to compare INSIGHTS opportunities with, three text-based prompts that were equivalent to the more structured INSIGHTS request and typical of user interactions with Notebook LM was input, to generate sets of 30 policy ideas, 30 business ideas and 30 technical design ideas in turn. The prompts were:

```
Develop thirty highly unusual [policy/business/technical design] ideas that support
seaside towns to regenerate by attracting new investment linked to new areas of
growth. Describe each idea with a short title and up to 100 words
```

Each was input to the web browser version of Notebook LM on 22$^{nd}$ and 23$^{rd}$ July 2025.

The second general-purpose tool was Open AI's ChatGPT-4o. Anecdotal evidence suggested that many innovation consultants use it to generate ideas for projects, hence the comparison. This time the comparison was with the public GPT4o model – no information assets were uploaded. The same prompts were input via a web browser on 22$^{nd}$ and 23$^{rd}$ July 2025 to generate sets of 30 policy ideas, 30 business ideas and 30 technical design ideas.

The two comparisons were undertaken to investigate the contribution of the computational model of creative outcomes to generate novel and useful innovation opportunities compared to outputs generated by commonly used LLMs. It explored the contributions of the **generative-rules** function to generate more novel opportunities and **classes-of-needs** function to generate more useful opportunities through two hypotheses:

H1: Opportunities generated by INSIGHTS were more creative, i.e., more novel and more useful than opportunities generated by Notebook LM for the same innovation consulting task

H2: Opportunities generated by INSIGHTS were more creative, i.e., more novel and more useful than opportunities generated by ChatGPT4-o for the same innovation consulting task

## 5.3 Comparing different INSIGHTS model functions

Different functions of the computational model were also compared using the novelty and usefulness ratings of opportunities generated with them. Hypotheses H3 and H4 explored whether the **atypical-opportunities** and **creative-qualities** functions increased novelty within opportunity spaces without reducing usefulness:

H3: Opportunities generated by INSIGHTS using the **atypical-opportunities** function were more novel and at least as useful than opportunities generated without that function, for the same innovation consulting task

H4: Opportunities generated by INSIGHTS using the **creative-qualities** function were more novel and at least as useful than opportunities generated without that function, for the same innovation consulting task

To investigate H3, the INSIGHTS baseline request was modified to generate opportunities that were very prototypical instead of highly unusual. For H4, the baseline request was modified to generate opportunities with one then two creative qualities instead of none. The contribution of the **pivot-opportunities** function was investigated using the novelty ratings of opportunities generated using the baseline request before and after the first pivot and second pivot:

H5: Opportunities generated by INSIGHTS using the **pivot-opportunities** function were more novel than opportunities generated without that function, for the same innovation consulting task

Figure 6 depicts the evaluation comparisons for the five hypotheses. Each compared novelty and usefulness ratings of opportunities generated by INSIGHTS computational model functions and the two LLM-based tools.

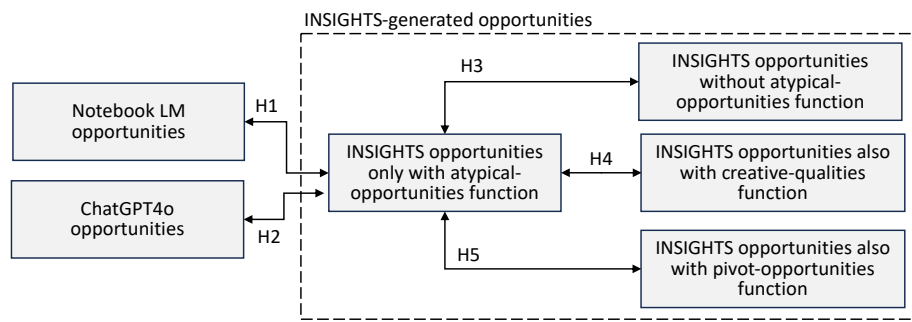

*Figure 6: Graphical depiction of the evaluation comparisons for each of the five hypotheses*

### 5.4 Rating the generated opportunities for novelty and usefulness

One method for rating opportunity novelty and usefulness is human expert judgment (e.g., [36] [37]). Although effective, it can be resource- and expertise-intensive and difficult to repeat, which can limit the scaling of any evaluation. This evaluation generated over 2000 opportunities to rate, so a more mechanized approach was adopted using customized prompts to GPT4o. Figure 7 presents the final prompt used to generate novelty and usefulness ratings for business innovation opportunities. Slightly different versions of the prompt were used to evaluate opportunities that were policy opportunities and technical design opportunities. The prompt generated two integers for each opportunity: 1) its novelty rating on a scale of 1-7, and 2) its usefulness rating on a scale of 1-7. All ratings were computed using ChatGPT4o between 28[th] and 30[th] July 2025. All generated novelty and usefulness ratings were analyzed using two statistical techniques applicable to ratings data. The first, the Mann-Whitney test, was applied to compare unpaired and unequal sets of novelty and usefulness ratings. The second, the Kruskal-Wallis test, was applied to compare three unmatched groups.

> You are an expert evaluator of innovation opportunities that innovation consultants can implement on behalf of clients. Assess the relative novelty and usefulness of lists of opportunities, even outside your own domain, relevant to the specific client challenge. Be unbiased and analytical in your evaluation. The client's [policy/business/technical design] challenge is:
>
> Develop an innovative [policy/business/technical design] opportunity that support seaside towns to regenerate by attracting new investment connected with new areas of growth
>
> The opportunities are:
> [Opportunities list here]
>
> Examine each opportunity for its strengths and weaknesses, its relevance to the above challenge, how well it can overcome the challenge, how similar or different it is to previous solutions implemented to address the challenge, and how different it is to previous solutions implemented in the UK hospitality sector.
>
> Use the outcomes of these examinations to rate each policy opportunity in the list using two criteria:
>
> - The opportunity's novelty on a scale of 1 to 7, where 1 is a very commonplace opportunity for overcoming the client's challenge and 7 is a highly unusual, novel and rare opportunity for overcoming the client's challenge
> - The opportunity's usefulness on a scale of 1 to 7, where 1 is not at all useful for overcoming the client's challenge and 7 is extremely useful and effective for overcoming the client's challenge
>
> Each rating should be an integer that is 1, 2, 3, 4, 5, 6 or 7. Present the ratings in a table of two columns and 30 rows of the novelty ratings then the usefulness ratings. The table can be downloaded.

*Figure 7. The automated GPT to generate novel and useful ratings on 1-7 integer scales for input lists of opportunity titles and their descriptions.*

### 5.5. Evaluation results

The evaluation generated a total of 2070 policy, business and technical design opportunities for the example UK hospitality sector, 1890 of which were generated by INSIGHTS, 180 by Notebook LM and 180 by ChatGPT4o. Each opportunity was then rated for its novelty and usefulness, resulting in a total of 4140 ratings.

#### 5.5.1. Comparing INSIGHTS to Notebook LM and ChatGPT

Overall averages of the novelty and usefulness ratings of opportunities on 1-7 scales generated by INSIGHTS using the baseline request, and by Notebook LM and ChatGPT4o with the equivalent prompts are presented in Table 3. These averages revealed that INSIGHTS generated policy, business and technical design opportunities that were both relatively novel (6+ out of 7) and useful (5.8+ out of 7).

|  | INSIGHTS opportunities | | Notebook LM opportunities | | ChatGPT4o opportunities | |
| --- | --- | --- | --- | --- | --- | --- |
|  | Novelty | Usefulness | Novelty | Usefulness | Novelty | Usefulness |
| Policy opportunities | 6.1 | 5.9 | 5.4 | 5.6 | 6.2 | 4.8 |
| Business opportunities | 6.2 | 5.9 | 5.5 | 5.3 | 5.9 | 4.6 |
| Technical design opportunities | 6.2 | 5.8 | 5.5 | 5.4 | 6.2 | 4.3 |

*Table 3. Averages of the novelty and usefulness ratings of opportunities on 1-7 scales generated by INSIGHTS, ChatGPT and Notebook LM.*

Comparing opportunities generated by INSIGHTS with the baseline request and Notebook LM, Mann-Whitney tests revealed significant differences in their novelty ratings (z=4.6448, p<.00001) and usefulness ratings (z=4.8593, p<.00001). Analysing further, Mann-Whitney tests revealed significant differences between novelty ratings for policy opportunities (z=2.0033, p=.02276), business opportunities (z=3.0086, p=0.0013) and technical design opportunities (z=2.9938, p=.0014). Mann-Whitney tests also revealed significant differences between usefulness ratings for business opportunities (z=2.4468, p=.0071) and technical design opportunities (z=2.3216, p=.0102) but not between usefulness ratings for policy opportunities (z=1.5893, p=.05592). Overall, the analysis revealed that INSIGHTS generated more creative, i.e., more novel and more useful business opportunities and technical design opportunities than Notebook

LM. Policy opportunities generated by INSIGHTS were more novel but not more useful than the opportunities generated by Notebook LM.

Comparing the opportunities generated by INSIGHTS and ChatGPT4o, Mann-Whitney tests revealed a significant difference in their opportunity usefulness ratings (z=8.538, p<.00001) but no significant differences in opportunity novelty ratings (z=0.2775, p=.3897). Analysing further, Mann-Whitney tests revealed significant differences in the usefulness ratings for policy opportunities (z=3.3191, p<.0004), business opportunities (z=4.4871, p<.0001) and technical design opportunities (z=5.5442, p<.0001). Overall, the analysis revealed that ChatGPT4o generated opportunities of different types that were as novel as INSIGHTS with the baseline request, but these opportunities were significantly less useful and hence creative than those generated by INSIGHTS.

### 5.5.2. Comparing different functions of the INSIGHTS computational model

Averages of the novelty and usefulness ratings of opportunities on 1-7 scales generated by the different functions of the INSIGHTS computational model are presented in Table 4.

|  | Columns A | | Columns B | | Columns C | | Columns D | |
|---|---|---|---|---|---|---|---|---|
|  | Opportunities are highly unusual, no creative qualities [baseline request] | | Opportunities are highly prototypical, no creative qualities | | Opportunities are highly unusual, one creative quality | | Opportunities are highly unusual, two creative qualities | |
|  | Usefulness | Novelty | Usefulness | Novelty | Usefulness | Novelty | Usefulness | Novelty |
| Policy | 6.1 | 6.0 | 5.8 | 4.6 | 5.9 | 6.3 | 6.1 | 6.0 |
| Business | 5.9 | 5.9 | 6.0 | 5.1 | 5.7 | 5.8 | 5.8 | 5.9 |
| Tech design | 5.8 | 6.2 | 5.8 | 5.2 | 5.9 | 6.0 | 5.9 | 6.0 |

*Table 4. Averages of usefulness and novelty ratings calculated for policy, business and technical design opportunities generated by computational model functions of INSIGHTS. Columns A present rating averages for the baseline request for opportunities generated with the **atypical-opportunities** function. Columns B present average ratings without the **atypical-opportunities** function switched on. Columns C and D present average ratings with the **creative-qualities** function switched on using one quality and two qualities.*

**Atypical-opportunities**: Mann-Whitney tests indicated that opportunities generated with topic terms that were less prototypical of selected opportunity spaces had significantly higher novelty ratings than opportunities generated with topic terms that were very prototypical of the same spaces (z=-11.48331, p <.00001). Analysing further, this significant difference existed for policy opportunities (z=-8.7416, p<.0001), business opportunities (z=-4.9351, p<.00001), and technical design opportunities (z=-6.5086, p< .00001). At the same time, the usefulness ratings for the same sets of opportunities were not significantly different for policy opportunities (z=-1.2860, p=.09853), business opportunities (z=1.0299, p=.1515), and technical design opportunities, z=.66494, p=.2579). Overall, the **atypical opportunities** function generated opportunities that were more novel and just as useful as prototypical opportunities in one opportunity space.

**Creative-qualities**: Kruskal-Wallis tests revealed no significant differences in the novelty ratings of opportunities generated using requests with no creative qualities, one creative quality and two creative qualities for business opportunities (H=0.5765 (2, N=270), p=.7496) and technical design opportunities (H=2.6307 (2, N=270), p=.2684), but it did reveal a significant difference in novelty ratings for policy opportunities (H=10.9454 (2, N=270), p=.0042). Kruskal-Wallis tests also revealed no significant difference in usefulness ratings generated with no creative qualities, one creative quality and two creative qualities for policy opportunities (H=0.7266 (2,N=270), p=.6954), business opportunities (H=0.4538 (2,N=270), p=.797) and technical design opportunities (H=2.4114 (2,N=270), p=.2995). Overall, opportunities generated using the INSIGHTS **creative-qualities** function were not more novel than opportunities generated without it, although policy opportunities generated using the **creative-qualities** function were more novel.

**Pivot-opportunities**: Average ratings for all (not just baseline) generated opportunities before and after the first pivot and after the second pivot are presented in Table 5. A Mann-Whitney test revealed a significant difference in novelty ratings for all opportunities generated before and after the first pivot ($z=-4.4309$, $p<.0001$). Tests also revealed these significant differences were present between policy opportunities ($z=-1.0423$, $p=.1492$), business opportunities ($z=-0.8191$, $p=.0019$) and technical design opportunities ($z=-3450$, $p=.0002$). Mann-Whitney tests also revealed no significant differences in usefulness ratings for opportunities generated before and after the first pivot for policy opportunities ($z=-0.0348$, $p=.488$), business opportunities ($z=1.3472$, $p=.0885$) and technical design ($z=1.1854$, $p=.117$) opportunities. Overall, applications of the pivot generative rules to opportunities generated from opportunity space topics increased the novelty of subsequent opportunities and did not impact the usefulness of these opportunities. Opportunities generated using the **pivot-opportunities** function were more novel and at least as useful than opportunities generated without it.

|  | Before first pivot | | After first pivot | | After second pivot | |
| --- | --- | --- | --- | --- | --- | --- |
|  | Usefulness | Novelty | Usefulness | Novelty | Usefulness | Novelty |
| Policy opportunities | 6.0 | 5.6 | 6.0 | 5.8 | 5.9 | 5.8 |
| Business opportunities | 5.9 | 5.5 | 5.8 | 5.9 | 5.8 | 5.7 |
| Technical design opportunities | 5.9 | 5.6 | 5.8 | 6.0 | 6.5 | 6.0 |

*Table 5. Average usefulness and novelty ratings for all generated (not just baseline) opportunities before and after first pivot, and after second pivot, using the **pivot-opportunities** function.*

By contrast, Mann-Whitney tests revealed no significant differences in novelty ratings for opportunities generated during the first and second pivots, for policy opportunities ($z=0.2073$, $p=.4168$), for business opportunities ($z=0.8191$, $p=.2061$) and for technical design ($z=-0.4509$, $p=.3264$) opportunities. This analysis revealed no further effect on opportunity novelty and usefulness from additional applications of the function.

Returning to the hypotheses, INSIGHTS with its baseline **atypical-opportunities** function generated more novel and useful opportunities than Notebook LM (H1 was accepted) and as novel but more useful opportunities than ChatGPT4o (H2 was rejected). The higher novelty ratings of the ChatGPT4o opportunities were perhaps surprising but offset by much lower usefulness ratings (even compared to Notebook LM), making the ChatGPT4o opportunities less creative. The **atypical-opportunities** function was necessary to generate the more novel opportunities (H3 was accepted), and **pivot-opportunities** function use further increased opportunity novelty ratings without reducing usefulness rating (H5 was accepted). By contrast, using the **creative-qualities** function did not generate opportunities with more novel ratings (H4 was rejected).

**5.6. Threats to the validity of evaluation results**

The evaluation results are subject to validity threats [38]. One threat to external validity, which limits our ability to generalize results, was the limited number of computational model settings applied to a single project. Therefore, we present the results as formative and leading to model improvements in a design science context [Wieringa 2014], to encourage others to access the INSIGHTS tool to undertake model evaluations.

Threats to construct validity limit the generality of the evaluation results to underpinning theory, i.e. the descriptive model of creative outcomes. The computational implementation required us to make many design decisions beyond its description, such as the numbers of opportunity spaces to generate or topic terms to extract. Each of these could have led to different evaluation outcomes. In mitigation, however, the implementation has evolved over time (e.g., 10+ versions of the generative rule prompts have been tested) to be more effective using feedback during informal evaluations with multiple organisations. This contributed to the ecological validity of the results, but can limit our claims to the current model implementation.

Different threats to the validity of our conclusions about the relations between uses of the different tools and reported outcomes were identified. An obvious one was the capability of ChatGPT4o to generate accurate opportunity evaluation ratings. Although evidence is growing for LLM use in research [39], both to generate synthetic research data (e.g., [40]) and support data coding (e.g., [41]), the authors were unaware of its use for rating creative outcomes, although a lightweight graph-based LLM framework for idea evaluation was reported [42] and research idea generation using LLMs was benchmarked [43]. One mitigating factor was the evaluation's comparison of different set of opportunities using the ratings generated using a single method over a two-day period. Rather than use ChatGPT4o to define opportunities that were novel or useful per se, it was used to explore relative differences between ratings generated using the same method, in the same two-day window.

Finally, threats to the evaluation's internal validity were influences that could have affected independent variables related to causality. One obvious source of an uncontrolled variable was the model's dependence on public LLMs (in our case GPT4o) to generate descriptions of opportunities and spaces. The profusion of new LLMs (e.g. GPT5) with different characteristics can affect model outcomes in difficult-to-predict ways. Again, one mitigation was that the computational model was compared to two other tools using the same or equivalent LLMs.

## 6. Conclusions and future work

Descriptive and computational versions of a new model of creative outcomes were developed and evaluated. A key evaluation finding was the contribution of spaces extracted from project information to generate useful as well as novel opportunities, consistent with [17] [18]. Alternative LLM applications such as ChatGPT4o and Notebook LM rely on run-time probabilistic manipulation of language tokens, even when the LLMs are generated from project-specific documents, which risks over-fitting when discovering clusters that equate to spaces of useful opportunities. The computational model's implementation of machine learning to generate clusters of topic terms associated with solutions classes [17] appears to have directed the generation of opportunities more likely to be useful, as well as provide consultants with more control, consistent with human-centred AI principles [6]. Opportunity spaces generated by the model appeared to define the boundaries to what was useful, and overcame the novelty-usefulness trade-off often observed in ideas generated from LLMs (e.g., [26]).

Furthermore, the model's **generative-rules** and **atypical-opportunities** functions applied to opportunity spaces generated more novel opportunities. Opportunities generated using lower-weighted topic terms were more novel but still useful – a finding that contrasted with idea generation using LLMs [4]. The model's current use of ten topic terms per space was reached via informal testing, but is ten terms the optimum? To find out, we are extending the **atypical-opportunities** function to generate more topic terms, to explore the boundaries of opportunity spaces that generate novel opportunities without losing usefulness.

The model's **pivot-opportunities** function also generated opportunities that were more novel, but subsequent pivots did not further increase novelty, and reasons for this were not clear. Therefore, we are currently exploring different manipulations of topic terms that provide the subject matter of new opportunities to increase novelty without impacting usefulness by expanding selected topic terms with less-common synonyms (e.g. *nature* to *bio-commons*) based query expansion techniques (e.g. [44] [45] [46]).

On the other hand, opportunities generated using the model's **creative-qualities** function were not more novel than opportunities generated without it (H4). The implemented creative qualities were developed to provide shortcuts for human creative thinking towards spaces of opportunities with these attributes [9] and add value when human cognitive resources were limited (e.g., [32]). However, the computational model used large computational resources to explore very large numbers of opportunities, which might have nullified their potential value to discover new spaces of opportunities.

In parallel, the computational model will undergo more evaluations by consultants using the INSIGHTS tool during innovation projects. E.g., a new INSIGHTS project has been set up with over 400 policy and research documents related to violence in society, to support researchers explore more policy innovations related to violence. These explorations will use new features such as interactive visualization of the opportunity spaces [47], evaluate interactive tool capabilities beyond the computational model [48], and will record the generated opportunities to analyze using experts for the relevant policy areas.